%% file: root.tex
\documentclass[letterpaper, 10 pt, conference]{ieeeconf}  

\IEEEoverridecommandlockouts                              

\title{Explaining Navigation Decisions: Stage-Aware Visual Attribution for Vision-Language Navigation}

\author{
Yiyuan Yao, Xinpeng Ding$^{*}$, Minglei Chen,
Fei Gao, Lihuo He,\\
Jie Yuan, Aijun Li, Nannan Wang, and Xinbo Gao%
\thanks{$^{*}$Corresponding author: Xinpeng Ding.}%
\thanks{Yiyuan Yao is with Xidian University,
Xi'an, China.}%
}

\usepackage{placeins}
\usepackage{amsmath,amssymb}
\usepackage{graphicx}
\usepackage{tikz}
\usetikzlibrary{arrows.meta,calc,fit,positioning}
\usepackage{booktabs}
\usepackage{multirow}
\usepackage[table]{xcolor}
\usepackage{subcaption}
\usepackage{float}
\usepackage{placeins}
\usepackage{microtype}

\def\ie{\textit{i.e.}}

\newcommand{\temp}{\textsc{TR}}
\newcommand{\spatial}{\textsc{SF}}
\usepackage{xcolor}
\definecolor{stageblue}{HTML}{315C69}
\definecolor{stagebrown}{HTML}{94745B}

\usetikzlibrary{arrows.meta,calc,fit,positioning}
\begin{document}

\maketitle
\thispagestyle{empty}
\pagestyle{empty}

\renewcommand{\paragraph}[1]{\vspace{0.05mm}\noindent\textbf{#1}}
\begin{abstract}
%
As Vision-Language Models (VLMs) increasingly enable robots to follow
language instructions, understanding their navigation decisions becomes
important for reliable deployment.
Existing explainability methods overlook a key challenge in
Vision-Language Navigation (VLN): evidence for the current action may be
distributed across historical and current observations and associated with
different instruction stages.
We identify \emph{instruction-stage insensitivity}, where existing methods
produce similar explanations despite changes across different navigation stages.

To address this issue, we propose a stage-aware attribution framework for VLN.
We first construct action-conditioned temporal instruction--visual
attributions across observations.
These attributions are projected into instruction space to estimate stage
associations, which guide the aggregation of relevant historical evidence for
refining the current explanation.
An instruction-guided spatial filter further suppresses isolated responses
while preserving compact, instruction-relevant regions.
Experiments across multiple VLN policies and benchmarks show improved
attribution quality over existing methods.

We further demonstrate that our method acts as a diagnostic tool for analyzing failure cases, and validate its applicability to real-world navigation on a Unitree G1 robot.
\end{abstract}

\input{Secs/1_introduction}

\input{Secs/2_related_work}
\input{Secs/3_method_v4}
\input{Secs/4_experiments_v2}

\input{Secs/5_conclusion}

\bibliographystyle{IEEEtran}
\bibliography{IEEEabrv,references}

\end{document}

%% file: Secs/1_introduction.tex
\section{INTRODUCTION}

Advances in Vision-Language Models (VLMs)~\cite{liu2023visual,li2023blip,bai2025qwen3,li2024llava} have accelerated progress in Vision-Language Navigation (VLN), enabling embodied agents to translate natural-language instructions and visual observations into navigation actions~\cite{cheng2024navila}.
Recent models~\cite{zhang2024navid,cheng2024navila,wei2025streamvln,wei2026ground} build on these capabilities to incorporate visual history into action prediction and connect multimodal understanding with robot execution, demonstrating instruction-following navigation in real-world environments~\cite{song2025towards}.
Despite these advances, explaining which visual evidence supports a particular navigation action and how it relates to the instruction remains underexplored.
Such understanding is especially important in safety-critical applications such as robotic guidance for visually impaired users, where navigation errors can compromise human safety~\cite{zhang2026safevla}.

While numerous explainability methods have been developed for visual recognition~\cite{selvaraju2017grad,chefer2021transformer} and vision-language generation~\cite{li2025token,bousselham2026dex}, explaining long-horizon navigation decisions remains challenging.
Methods such as TAM and DEX-AR address contextual interference and varying visual dependence in generated text, but are not specifically designed to explain embodied action decisions. Although emerging methods such as ISS have begun to address embodied action explainability through visual input perturbations~\cite{zhang2026embodied}, they do not explicitly associate the attributed visual evidence with instruction stages.
This limitation becomes particularly consequential in VLN, where both the visible scene and the relevant instruction stage change as the agent navigates through the environment.

\input{Items/fig_teaser}


As illustrated in Fig.~\ref{fig:teaser}~(b), consider the instruction
``walk past the pool table, then stop in front of the television.''
When explaining the stopping decision, evidence for the television may lie in
a recent observation rather than the current view, while earlier observations
still contain the pool table from a completed instruction stage.
This creates two challenges for VLN explanation:
\textbf{temporally distributed evidence}, where decision-relevant evidence is
spread across historical and current observations, and
\textbf{cross-stage interference}, where evidence from other instruction stages
can introduce irrelevant responses into the explanation.
Discarding history may therefore remove useful evidence, while retaining all
history without stage awareness can mix explanations from different subgoals.
Our analysis further reveals \textbf{\emph{instruction-stage insensitivity}}:
when the relevant instruction stage changes, existing methods often produce
only minor changes in their instruction-token and visual attribution patterns.
As shown in Fig.~\ref{fig:teaser}~(c), explanations remain highly similar
across decisions from different stages, failing to reflect the progression of
the navigation instruction.

We argue that explaining VLN decisions requires preserving the relationship between instruction stages and visual evidence across observations. To this end, we introduce a visual attribution framework tailored to long-horizon navigation. We first construct \emph{temporal instruction--visual attributions} conditioned on the current action, using attention weights and their gradients to retain instruction--patch associations in historical and current observations. Building on this representation, \emph{Instruction-Stage Attribution} projects each observation's attribution into instruction space, providing a soft descriptor of its associated instruction stage. \emph{Stage-Guided Temporal Refinement} \textbf{(\temp)} then aggregates historical instruction relevance according to its stage proximity to the current observation and uses this support to selectively refine the current-frame attribution. Finally, \emph{Instruction-Guided Spatial Filter} \textbf{(\spatial)} uses instruction--patch correspondence to moderate spatial suppression, reducing isolated responses while preserving compact regions with strong instruction associations. Together, these components expose historical evidence supporting the current action and refine its current-view explanation through stage-relevant temporal support and instruction-guided spatial processing.


Experimentally, our method consistently outperforms existing explainability methods on R2R~\cite{anderson2018vision} and RxR~\cite{ku2020room}, improving Precision by 34.58\% and 35.37\% and F1-Score by 25.72\% and 23.50\%, respectively, over DEX-AR~\cite{bousselham2026dex}, while maintaining competitive decision faithfulness.
Qualitative results further show more stage-consistent localization of instruction-relevant evidence. We also demonstrate its utility for diagnosing VLN failures and its applicability to real-world navigation on a Unitree G1 robot.

Our contributions are summarized as follows:
\begin{itemize}
    \item We study visual explainability for VLM-based VLN policies, with a particular focus on how explanations evolve across navigation stages. We identify and quantify \emph{instruction-stage insensitivity}, where
explanations change insufficiently as the relevant navigation stage progresses.
    \item We introduce temporal instruction--visual attribution to explain the current action across observations, and develop \temp~that uses stage-relevant historical support to refine current-frame attributions.
    \item We propose \spatial~that regulates spatial suppression with
    instruction--patch associations, preserving compact relevant evidence
    while reducing isolated responses.
    \item We demonstrate consistent gains across VLN policies and benchmarks,
    and validate the framework for failure diagnosis and real-world
    navigation on a Unitree G1.
\end{itemize}




%% file: Items/fig_teaser.tex
\begin{figure}[t]
    \centering
    \includegraphics[width=1\columnwidth]{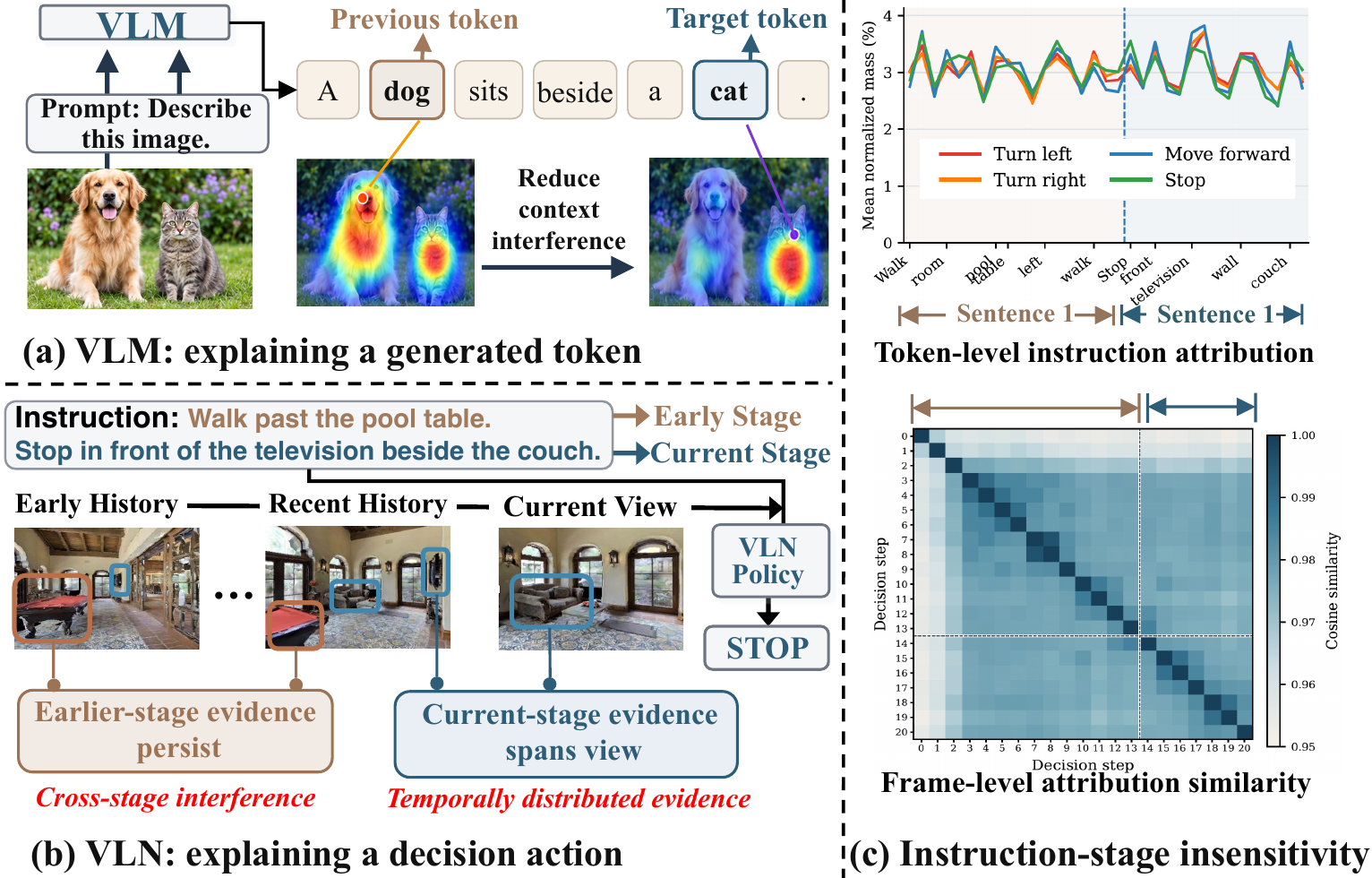}
    \vspace{-4mm}
   \caption{
   \textbf{Visual explanation challenges in VLN.} \textbf{(a)} Explaining a generated token involves localizing its visual evidence while reducing interference from preceding tokens. \textbf{(b)} Explaining a navigation action requires recovering 
   {\emph{temporally distributed evidence}} across historical and current observations, while suppressing 
   {\emph{cross-stage interference}} from evidence associated with other instruction stages. Colored boxes schematically indicate stage-related regions. \textbf{(c)} Existing methods~\cite{bousselham2026dex} are insensitive to instruction-stage transitions, exhibiting overlapping token attributions (\textbf{top}) and high cross-stage explanation similarity (\textbf{bottom}).
}
\label{fig:teaser}
\vspace{-3mm}
\end{figure}

%% file: Secs/2_related_work.tex
\section{Related Works}

\paragraph{Vision-Language Navigation.}
VLN has evolved from recurrent sequence-to-sequence
policies~\cite{anderson2018vision} to multimodal Transformers
that explicitly incorporate navigation history~\cite{chen2021history}.
Recent VLM-based approaches, including NaVid, NaVILA, and
Uni-NaVid~\cite{zhang2024navid,cheng2024navila,zhang2024uni},
integrate language understanding with visual observations
for generalizable action prediction.
Further advances explore efficient temporal memory
in StreamVLN and JanusVLN~\cite{wei2025streamvln,zeng2026janusvln},
dual-system planning and execution in DualVLN~\cite{wei2026ground},
and generalist navigation in LightNav-0~\cite{wang2026lightnav}.Recent studies further investigate progress-aware reasoning and selective history utilization for long-horizon navigation~\cite{guo2026awarevln}~\cite{wang2026progress}~\cite{xue2026profocus}.
These studies primarily advance navigation capability and efficiency.
Our work aims to explain how visual evidence across observations relates to instruction stages and supports the current action.

\paragraph{Visual explainability in vision-language models.}
Visual explainability has been widely studied for CNNs, including
Grad-CAM~\cite{selvaraju2017grad}, Guided Backpropagation~\cite{springenberg2014striving},
and Integrated Gradients~\cite{sundararajan2017axiomatic}.
With the rise of VLMs, explanation methods have been extended to multimodal
Transformers and CLIP-like models~\cite{gandelsman2024interpreting,bousselham2025legrad},
using gradient-attention attribution~\cite{chefer2021transformer,barkan2023visual}
or model-agnostic perturbation~\cite{petsiuk2018rise}.
More recent methods target autoregressive VLMs: TAM~\cite{li2025token}
reduces interference from preceding context tokens for token-specific visual
explanations, while DEX-AR~\cite{bousselham2026dex} combines layer-wise
attention gradients with dynamic head and token filtering for per-token and
sequence-level attribution.
However, these methods primarily explain semantic predictions or generated
tokens rather than embodied actions.

\paragraph{Visual explainability in embodied models.} Visual explainability remains relatively underexplored in embodied models. Recent work has also explored capability-level failure attribution for VLN agents~\cite{chen2026did}, aiming to localize failures to perception,
memory, planning, or decision-making components.ISS~\cite{zhang2026embodied} estimates visual importance in vision-language-action models by repeatedly masking image regions with blurred backgrounds and aggregating the resulting action deviations. However, these perturbations require repeated policy evaluations, making the approach costly to extend to VLN, where changing viewpoints and accumulated visual history introduce evidence across multiple observations at each step. Our method instead constructs action-conditioned instruction--visual attributions across observations and uses instruction-stage associations to refine the visual explanation without repeated input masking.

%% file: Secs/3_method_v4.tex
\section{Method}
\label{sec:method}

\input{Items/fig_framework}

\subsection{Vision-Language Navigation Preliminaries}
\label{sec:vln_preliminaries}

Let $X=\{x_i\}_{i=1}^{L}$ denote a navigation instruction.
At step $T$, given the current observation $V_T$ and history
$H_T=\{V_t\}_{t=1}^{T-1}$, a navigation policy $\pi_\theta$ predicts
\begin{equation}
    a_T=\pi_{\theta}(X,V_T,H_T).
\end{equation}

Conventional VLM attribution explains a target output $y$ by identifying
supporting visual evidence under textual context $X$:
\begin{equation}
    \text{VLM:}\qquad
    V \xrightarrow{\text{conditioned on }X} y.
\end{equation}
In VLN, however, evidence for the current action may span historical and
current observations and correspond to different parts of the instruction:
\begin{equation}
    \text{VLN:}\qquad
    \underbrace{\{V_t\}_{t=1}^{T}\leftrightarrow X}
    _{\text{temporal instruction--visual associations}}
    \longrightarrow a_T.
    \label{eq:vln_explanation_flow}
\end{equation}
This relation describes the explanation scope rather than the policy's
internal computation.

Because observations may contain evidence from different instruction stages,
direct aggregation can introduce \textbf{\emph{cross-stage interference}}
into the explanation of $a_T$.
Our framework addresses this by constructing temporal instruction--visual
attributions (Sec.~\ref{sec:temporal_attribution}), estimating their
instruction-stage relevance (Sec.~\ref{sec:instruction_stage}), refining
current attribution with stage-aligned historical support
(Sec.~\ref{sec:temporal_refinement}), and applying instruction-guided spatial
filtering to preserve compact relevant evidence
(Sec.~\ref{sec:spatial_filter}).

\subsection{Temporal Instruction--Visual Attribution}
\label{sec:temporal_attribution}

To distinguish evidence from different instruction stages, we retain both
its instruction association and temporal location.
For each observation $V_t$, we therefore construct an instruction--patch
attribution conditioned on the selected action $a_T$.

Let $\mathbf{A}^{\mathrm{full}}_{l,h}\in\mathbb{R}^{S\times S}$ denote
the self-attention matrix over the multimodal sequence at layer $l$ and
head $h$~\cite{bousselham2026dex}.
Let $\mathcal{I}$ and $\mathcal{V}_t$ denote the instruction-token and
visual-patch positions, respectively.
We extract
\begin{equation}
    \mathbf{A}_{l,h,t}
    =
    \mathbf{A}^{\mathrm{full}}_{l,h}
    [\mathcal{I},\mathcal{V}_t]
    \in \mathbb{R}^{L\times N_t},
\end{equation}
where $\mathbf{A}_{l,h,t}(i,j)$ is the attention from instruction token
$x_i$ to visual patch $v_{t,j}$.

Let $F$ be the logit of the selected action $a_T$.
Since attention magnitude alone does not measure action relevance, we combine
attention with its gradient and aggregate positive contributions:
\begin{equation}
\resizebox{0.90\columnwidth}{!}{$\displaystyle
\mathbf{M}_t(i,j)=
\sum_{l\in\mathcal{L}}\sum_h \alpha_{l,h}
\left[
\mathbf{A}_{l,h,t}(i,j)
\frac{\partial F}
{\partial \mathbf{A}_{l,h,t}(i,j)}
\right]_{+}.
$}
\label{eq:temporal_attribution}
\end{equation}
Here, $[x]_+=\max(x,0)$, $\mathcal{L}$ denotes the inspected layers, and
$\alpha_{l,h}$ is a normalized sensitivity weight computed from the maximum
absolute gradient within each instruction--visual attention submatrix.
Gradients through multimodal prefill attention are retained using a fully
differentiable forward pass.

Each $\mathbf{M}_t\in\mathbb{R}_{\geq0}^{L\times N_t}$ assigns an
action-conditioned attribution to every instruction--patch pair.
We collect these matrices in observation order to form the
\textbf{temporal instruction--visual attribution sequence}:
\begin{equation}
    \mathcal{M}
    = \bigl(
    \mathbf{M}_1,\ldots,\mathbf{M}_{T-1},\mathbf{M}_T
    \bigr).
    \label{eq:temporal_attribution_sequence}
\end{equation}
Here, $\mathbf{M}_T$ corresponds to the current observation,
and the preceding matrices correspond to historical observations.
By retaining the instruction dimension within each frame,
$\mathcal{M}$ allows us to characterize the instruction stages
associated with visual evidence before aggregating it across time.

\subsection{Instruction-Stage Attribution}
\label{sec:instruction_stage}

\paragraph{Instruction-space projection.}
We project each $\mathbf{M}_t$ into instruction space by marginalizing over
visual patches:
\begin{equation}
\mathbf{p}_t(i)
=
\frac{\sum_j \mathbf{M}_t(i,j)}
{\sum_{i',j}\mathbf{M}_t(i',j)+\epsilon},
\qquad 1\leq t\leq T,
\label{eq:instruction_projection}
\end{equation}
where $\mathbf{p}_t$ describes how action-conditioned visual evidence is
distributed over instruction tokens.

\paragraph{Instruction-stage estimation.}
We summarize this distribution by its normalized positional expectation:
\begin{equation}
\mu_t
=
\sum_{i=1}^{L}
\frac{i-1}{L-1}\mathbf{p}_t(i).
\label{eq:stage_estimation}
\end{equation}
The scalar $\mu_t$ provides a soft descriptor of the instruction stage
associated with evidence in $V_t$, rather than task completion.
We use $\mu_T$ as the current-stage reference while retaining
$\mathbf{p}_t$ for token-level instruction support.

\subsection{Stage-Guided Temporal Refinement}
\label{sec:temporal_refinement}

Historical observations may support the current action while belonging to
different instruction stages.
We therefore use stage correspondence to select relevant historical support
for refining the current attribution.

\paragraph{Stage weighting.}
For each historical observation $t<T$, we compute
\begin{equation}
    \gamma_t
    =
    \frac{
        \exp\!\left(-|\mu_T-\mu_t|/\tau_H\right)
    }{
        \sum_{s=1}^{T-1}
        \exp\!\left(-|\mu_T-\mu_s|/\tau_H\right)
    },
    \label{eq:temporal_weights}
\end{equation}
where $\tau_H>0$ controls sensitivity to stage differences.
Observations closer to the current instruction stage receive larger weights.

\paragraph{Temporal refinement.}
We aggregate stage-aligned historical support as
\begin{equation}
    \bar{\mathbf{p}}_T(i)
    =
    \sum_{t=1}^{T-1}\gamma_t\mathbf{p}_t(i),
    \label{eq:historical_support}
\end{equation}
and use it to reweight the current-frame attribution:
\begin{equation}
    \widetilde{\mathbf{M}}_T(i,j)
    =
    \mathbf{M}_T(i,j)
    \left[
        \delta+(1-\delta)\bar{\mathbf{p}}_T(i)
    \right].
    \label{eq:temporal_refinement}
\end{equation}
Here, $\delta\in[0,1]$ sets the minimum retained response, preventing weak
historical support from suppressing newly observed evidence.
This refinement reduces cross-stage interference along the instruction
dimension while preserving the spatial localization in $\mathbf{M}_T$.
Historical attribution maps remain available in $\mathcal{M}$ for identifying
earlier evidence supporting the same action.

\subsection{Instruction-Guided Spatial Filter}
\label{sec:spatial_filter}

Temporal refinement reduces cross-stage interference, but isolated spatial
responses may remain.
Rank-Gaussian filtering~\cite{li2025token} can suppress such responses, yet
may also remove compact task-relevant regions.
We therefore regulate spatial suppression using instruction--patch
associations.

\paragraph{Visual map.}
We first obtain a patch-level map:
\begin{equation}
    \mathbf{g}_T(j)
    =
    \sum_{i=1}^{L}\widetilde{\mathbf{M}}_T(i,j).
    \label{eq:visual_readout}
\end{equation}
While $\mathbf{g}_T$ gives the total attribution of each patch, the original
instruction--patch associations are retained in
$\widetilde{\mathbf{M}}_T$.

\paragraph{Instruction-based preservation.}
We characterize each association from both patch and token perspectives:
\begin{equation}
\begin{aligned}
    q_T(i\mid j)
    &=
    \frac{\widetilde{\mathbf{M}}_T(i,j)}
         {\sum_{i'}\widetilde{\mathbf{M}}_T(i',j)+\epsilon},\\
    q_T(j\mid i)
    &=
    \frac{\widetilde{\mathbf{M}}_T(i,j)}
         {\sum_{j'}\widetilde{\mathbf{M}}_T(i,j')+\epsilon}.
\end{aligned}
\label{eq:bidirectional_grounding}
\end{equation}
Here, $q_T(i\mid j)$ measures how strongly patch $j$ is attributed to token $i$, while $q_T(j\mid i)$ measures how strongly token $i$ is localized to patch $j$. Their geometric mean favors associations that are concentrated in both directions.
We define the preservation score of each patch by its strongest association with an instruction token:
\begin{equation}
\mathbf{s}_T(j)
=
\max_i
\sqrt{q_T(i\mid j)\,q_T(j\mid i)}.
\label{eq:preservation_score}
\end{equation}
A high $\mathbf{s}_T(j)$ indicates a strong, compact instruction--patch
association and therefore greater resistance to spatial suppression.

\paragraph{Controlled spatial suppression.}
Let $\mathcal{D}_k$ denote the rank-Gaussian filter with neighborhood
parameter $k$.
We modulate its proposed suppression using $\mathbf{s}_T$:
\begin{equation}
\resizebox{0.98\columnwidth}{!}{$\displaystyle
    \mathbf{g}_T^{*}(j)
    =
    \mathbf{g}_T(j)
    -
    \bigl(1-\mathbf{s}_T(j)\bigr)
    \left[
        \mathbf{g}_T(j)-\mathcal{D}_k(\mathbf{g}_T)(j)
    \right]_{+}.
$}
\label{eq:spatial_filter}
\end{equation}
Strongly grounded patches are preserved, whereas weakly associated responses
receive stronger suppression.
Thus, the filter removes isolated noise without discarding compact
instruction-relevant evidence.

%% file: Items/fig_framework.tex
\begin{figure*}[!t]
    \centering
    \includegraphics[width=2\columnwidth]{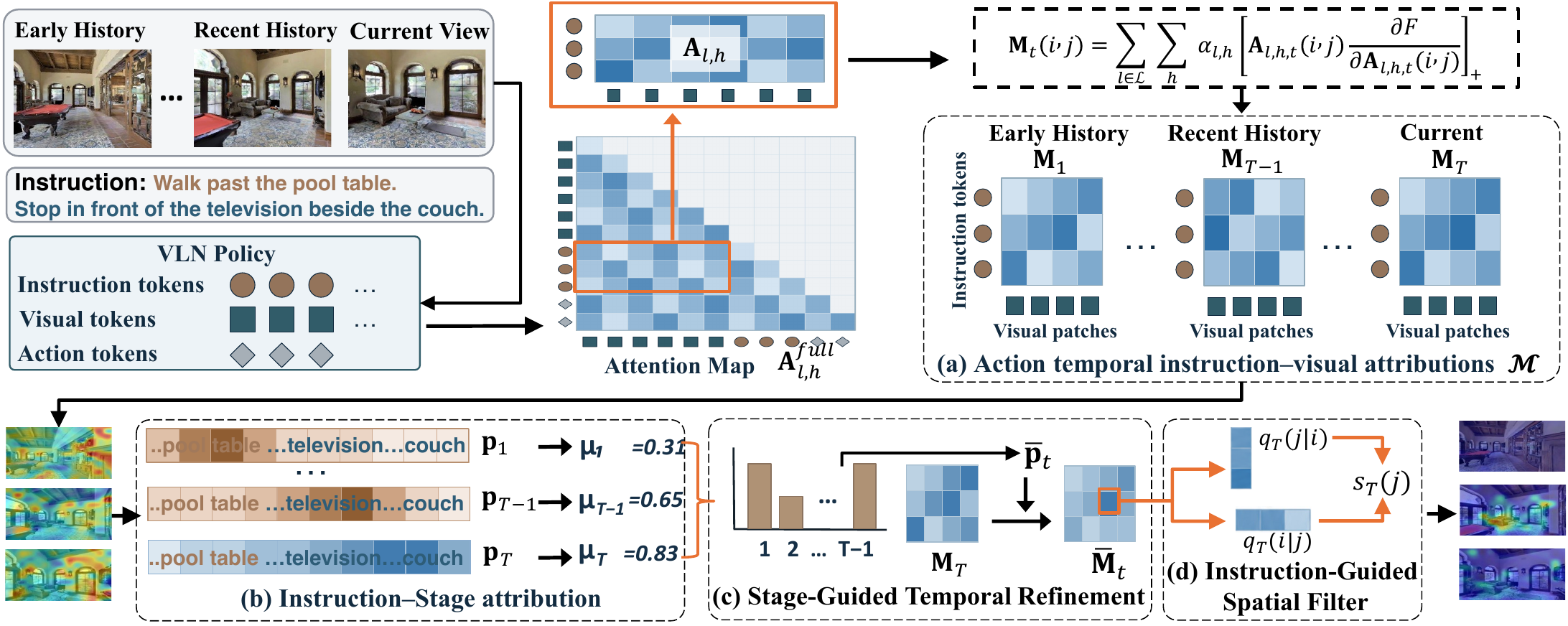}
    \vspace{-2mm}
   \caption{
\textbf{Overview of our framework.}
\textbf{(a)} We construct action-conditioned temporal instruction--visual attributions across historical and current observations.
\textbf{(b)} We project each attribution into instruction space to estimate its associated instruction stage.
\textbf{(c)} We weight historical observations by their stage correspondence with the current view and aggregate the resulting support to refine the current attribution.
\textbf{(d)} We further use instruction--patch associations to suppress spatial noise while preserving decision-relevant visual evidence.
}
\label{fig:framework}
\vspace{-4mm}
\end{figure*}

%% file: Secs/4_experiments_v2.tex
\section{Experiments}
\label{sec:experiments}

\input{Items/fig_sota}

\subsection{Experimental Setup}
\label{sec:experimental_setup}

\paragraph{Dataset.}
We evaluate our method on R2R~\cite{anderson2018vision} and
RxR~\cite{ku2020room} under the Vision-and-Language Navigation in
Continuous Environments (VLN-CE) setting~\cite{krantz2020beyond}.
Both benchmarks are instantiated in Habitat, where agents follow
natural-language instructions from RGB observations in continuous
3D environments. We conduct quantitative evaluations on both
benchmarks and use the available semantic scene annotations to
evaluate instruction-aware visual grounding.

\paragraph{Implementation Details.}
We evaluate on NaVILA~\cite{cheng2024navila}, NaVid~\cite{zhang2024navid},
and StereoNav~\cite{wang2026limits} under VLN-CE.
For action-conditioned attribution, we use language-model layers 8--15
with all attention heads and gradient-based sensitivity weighting.
We set $\tau_H=0.25$, $\delta=0.7$, and $k=3$.
Input resolutions are $384\times384$, $224\times224$, and $448\times448$
for NaVILA, NaVid, and StereoNav, respectively.
At each decision step, explanations are computed over the current view
and all valid model-selected historical observations.
\textbf{All baselines are adapted to the same selected action and evaluated on the identical policy inputs and observations.}



\paragraph{Metrics.}
We evaluate explanation quality from two complementary perspectives.


\textbf{Decision Faithfulness.}
A faithful explanation should identify visual evidence that actually supports
the policy's selected navigation action. We therefore introduce an
input-perturbation-based \textit{Action-Space Delta-NLL} metric that measures
how strongly the predicted action depends on the highlighted regions.
Given the selected action $a^\ast$, we mask the salient regions and compute
the increase in its negative log-likelihood:
\begin{equation}
    \Delta \mathrm{NLL}
    =
    \log q_{\mathrm{clean}}(a^\ast)
    -
    \log q_{\mathrm{masked}}(a^\ast).
\end{equation}
Since masking itself may perturb the policy, we subtract the average effect
of $R$ random masks:
\begin{equation}
    \Delta \mathrm{NLL}_{\mathrm{corr}}
    =
    \Delta \mathrm{NLL}_{\mathrm{salient}}
    -
    \frac{1}{R}
    \sum_{r=1}^{R}
    \Delta \mathrm{NLL}_{\mathrm{random},r}.
\end{equation}
A larger $\Delta \mathrm{NLL}_{\mathrm{corr}}$ indicates that the highlighted
regions contribute more specifically to the current action decision, and thus
correspond to a more faithful explanation.

\input{Items/tab_sota}

\textbf{Semantic Alignment.}
We evaluate whether the highlighted regions correspond to
instruction-relevant visual concepts. For each observation, semantic
scene annotations are converted into binary masks $\{M_l\}$ for
individual categories. A pretrained CLIP text
encoder~\cite{radford2021learning} is used to estimate the semantic
relevance $w_l$ between the navigation instruction and category $l$.
We then construct an instruction-conditioned soft semantic target:
\begin{equation}
    G_i
    =
    \max_l \left( w_l M_l(i) \right).
\end{equation}
The explanation heatmap $H$ is binarized using OTSU thresholding:
\begin{equation}
    P_i
    =
    \mathbf{1}[H_i > t^\ast].
\end{equation}
Following the spatial evaluation protocol of TAM~\cite{li2025token},
semantic alignment is measured as
\begin{equation}
    \mathrm{Precision}
    =
    \frac{
        \sum_i P_i G_i
    }{
        \sum_i P_i + \epsilon
    }.
\end{equation}
A higher Precision indicates stronger alignment with
instruction-relevant semantic regions.

\textbf{Overall Score.}
We additionally report a harmonic-mean score that jointly considers
decision faithfulness and semantic alignment:
\begin{equation}
    \mathrm{F1\text{-}Score}
    =
    \frac{
        2\cdot
        \Delta\mathrm{NLL}_{\mathrm{corr}}
        \cdot
        \mathrm{Precision}
    }{
        \Delta\mathrm{NLL}_{\mathrm{corr}}
        +
        \mathrm{Precision}
        +
        \epsilon
    }.
\end{equation}
A higher score indicates better overall explanation quality.

\input{Items/tab_generalization}

\subsection{Comparison with State-of-the-Art Methods}
\label{sec:sota_comparison}

\paragraph{Quantitative results.}
As shown in Table~\ref{tab:explanation_results}, our method achieves the
best overall F1-Score on both R2R and RxR. The gain mainly comes from
substantially improved semantic alignment, with Precision increasing from
0.3591 to 0.4370 on R2R and from 0.2862 to 0.4241 on RxR over the strongest
baselines.
Meanwhile, our method also achieves the strongest decision faithfulness.
On R2R, it obtains the highest $\Delta\mathrm{NLL}_{\mathrm{corr}}$
(0.3198), while simultaneously achieving substantially better semantic
alignment in terms of Precision and F1-Score. On RxR, it
also obtains the highest faithfulness score (0.2846). Overall, these results
show that our method better identifies instruction-relevant visual evidence
without sacrificing fidelity to the navigation decision.

\input{Items/fig_ablation_TR}
\input{Items/fig_ablation_SF}

\input{Items/tab_ablation}

\input{Items/fig_statics}

\input{Items/fig_real_world}

\input{Items/fig_failure_analyze}

\paragraph{Qualitative results.}
Fig.~\ref{fig:sota} further support the effect of our method. Across different navigation stages, our explanations consistently shift toward the visual evidence associated with the active sub-instruction, such as the staircase region in Step 1, the plant and hallway in Step 7, and the couch/fireplace area in Step 21.
Moreover, our method preserves relevant evidence across historical and current observations, reflecting the temporally distributed nature of VLN decision making. In contrast, DEX-AR~\cite{bousselham2026dex}~produces highly diffuse and noisy activations, while GradCAM~\cite{selvaraju2017grad} and TAM~\cite{li2025token} often highlight broad scene regions without clearly adapting to the current instruction stage. Together, the quantitative and qualitative results demonstrate that our method produces more stage-consistent, semantically aligned, and decision-faithful explanations for VLN policies.

\paragraph{Generalization across VLN policies.}
As shown in Table~\ref{tab:agent_results}, our method consistently
outperforms DEX-AR across NaVid, NaVILA, and StereoNav on both R2R and RxR.
The improvements are especially strong in Precision, reaching up to
34.58\% on R2R and 42.03\% on RxR.
This consistent gain across different policies and datasets indicates that
our stage-aware explanation framework generalizes beyond a specific VLN
architecture.



\subsection{Ablation Study}
\label{sec:ablation}

We ablate the two key components of our framework, stage-guided temporal refinement (\temp) and instruction-guided spatial filtering (\spatial), to evaluate their individual contributions. Table~\ref{tab:ablation} reports the results of progressively enabling each module.

\paragraph{Effect of \temp.}
As reported in Table~\ref{tab:ablation}, enabling \temp~alone improves
Precision from 0.4180 to 0.4328 and F1-Score from 0.3252 to 0.3386,
indicating better alignment with instruction-relevant evidence.
Figure~\ref{fig:temporal_refinement_ablation} further shows that \temp~reduces temporal cross-stage interference and produces cleaner, more focused
current-frame attributions. After refinement, irrelevant responses from other
instruction stages are suppressed, while evidence relevant to the current
navigation decision is enhanced.


\paragraph{Effect of \spatial.}
Table~\ref{tab:ablation} shows that enabling \spatial~alone yields a larger
gain in decision faithfulness, increasing $\Delta\mathrm{NLL}$ from 0.2661
to 0.3082 and F1-Score from 0.3252 to 0.3553.
Fig.~\ref{fig:semantic_filter_ablation} visualizes the effect of SF.
The base rank filter~\cite{li2025token} removes isolated responses but may also suppress compact instruction-relevant evidence,~\ie, ``the picture on the wall". Incorporating instruction--patch associations improves target preservation, and the bidirectional variant further reduces irrelevant activations while retaining both the dining-table and wall-picture regions.

\paragraph{Instruction-Stage Sensitivity Analysis.}
We further quantify whether the explanations respond to changes in the
instruction stage. For each trajectory, we normalize the instruction-token
attribution at each decision step and compute the Jensen--Shannon divergence
(JSD) between different action decisions.
For visual attribution, we compute the cosine similarity between flattened
heatmaps across different stages.
To measure spatial concentration, we stack the attribution values from the
eight evaluated observations and report the fraction of total attribution
mass contained in the top 10\% responses.
All metrics are averaged over the evaluation samples.
Figure~\ref{fig:statics} further quantifies instruction-stage sensitivity.
Compared with the baseline, our method produces a substantially higher
instruction-attribution JSD (0.33 vs.\ 0.02) and lower cross-stage visual
similarity (57.4\% vs.\ 96.8\%), showing that both instruction and visual
attributions change more clearly across navigation stages.
Meanwhile, the higher top-10\% attribution mass (43.4\% vs.\ 13.5\%) indicates more concentrated visual explanations.
Together, these results confirm that our method effectively alleviates
instruction-stage insensitivity.

\subsection{Real-World Evaluation}
\label{sec:real_world}
We further deploy our framework on a Unitree G1 humanoid robot.
Figure~\ref{fig:unitree} shows two representative navigation trajectories,
pairing egocentric action-conditioned attributions with third-person execution views.
Across different decision steps, the explanations consistently highlight visual evidence
relevant to the current instruction, demonstrating that the proposed attribution framework
transfers from simulation to real-world navigation.

\subsection{Attribution as a Diagnostic Tool}
\label{sec:diagnostic_tool}

We further use our method to diagnose failed VLN decisions.
Figure~\ref{fig:failure} shows three representative cases where visual and
instruction-token attributions reveal mismatches between the model's selected action and the instruction requirement:

\textbf{(a) Premature subgoal execution.}
The model should pass through the right doorway before continuing across the
hallway, but instead turns left.
The attribution shifts toward the later hallway clause, suggesting premature
progression to a subsequent subgoal.
\textbf{(b) Action-phrase neglect.}
The instruction requires passing the stove while entering the dining room.
Although the stove is correctly localized, token attribution emphasizes
``stove'' over ``passing by,'' and the model turns toward it.
This indicates correct landmark grounding but insufficient attention to the
action phrase.
\textbf{(c) Partial instruction grounding.}
The model should turn right at the living room and enter, but moves forward.
Attribution remains concentrated on ``kitchen'' from the preceding clause,
with weaker response to the turning instruction, indicating incomplete grounding of the current step.

These examples demonstrate that our attributions can expose stage mismatches,
action-phrase neglect, and incomplete grounding underlying erroneous VLN
decisions.

%% file: Items/fig_sota.tex
\begin{figure*}[!t]
    \centering
    \includegraphics[width=2\columnwidth]{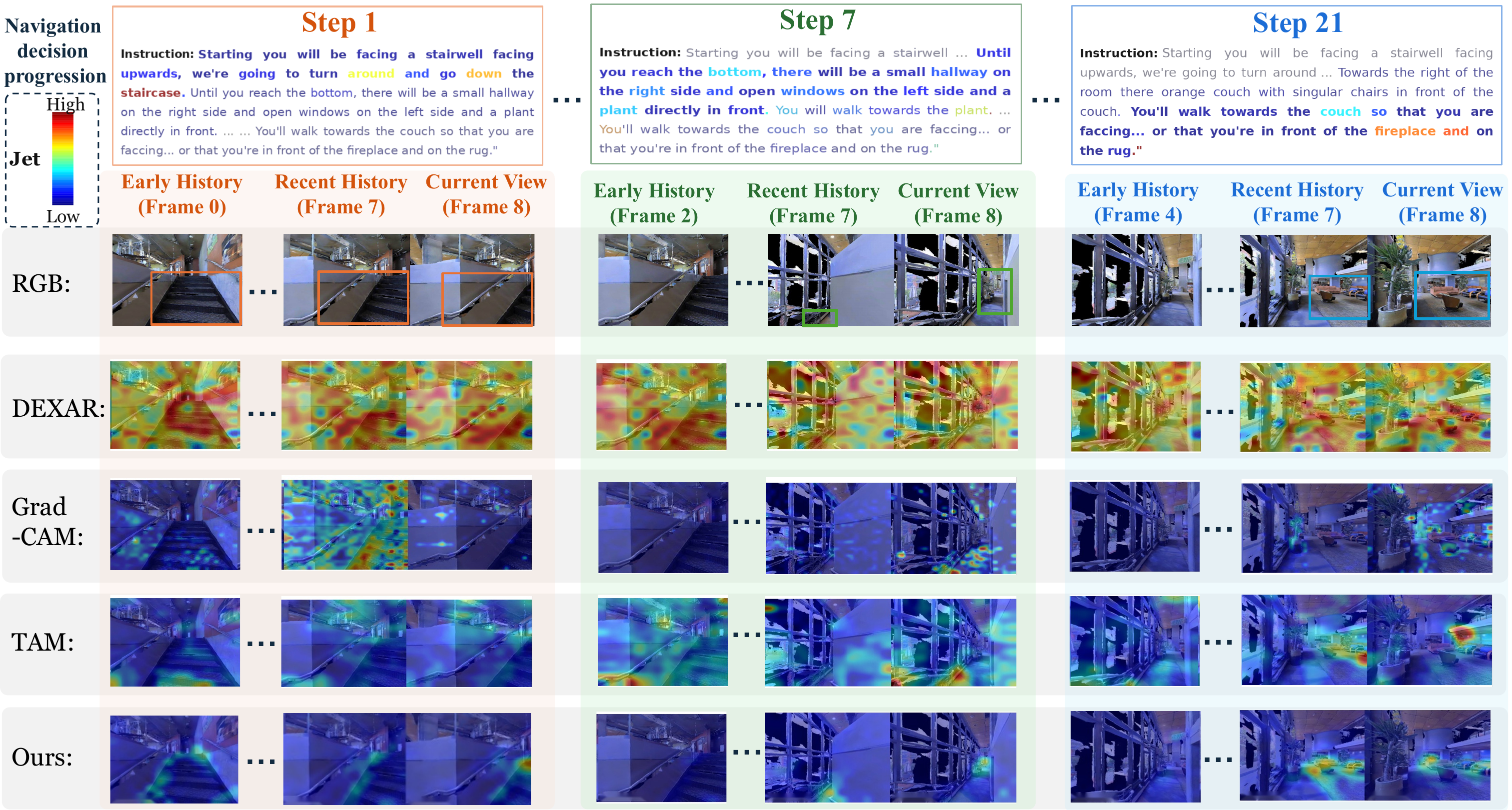}
   \caption{
   \textbf{Qualitative comparison with state-of-the-art methods across navigation stages.}
The same instruction is used throughout navigation, while the highlighted tokens are automatically induced by the action-conditioned instruction attribution at each decision step.
For Steps 1, 7, and 21, we visualize early-history, recent-history, and current observations together with their attribution maps.
Compared with existing methods, our explanations better follow the evolving instruction stage while retaining relevant evidence across historical and current views.
}
\label{fig:sota}
\vspace{-3mm}
\end{figure*}

%% file: Items/tab_sota.tex
\begin{table}[t]
\centering
\caption{\textbf{Quantitative comparison on R2R and RxR using NaVILA under the VLN-CE setting.}
We report decision faithfulness ($\Delta\mathrm{NLL}_{\mathrm{corr}}$),
semantic alignment (Precision), and F1-Score.
Higher is better for all metrics.}
\label{tab:explanation_results}
\resizebox{\linewidth}{!}{
\begin{tabular}{c|c|c|c c c}
\toprule
Dataset & Method & Type
& $\Delta$NLL(\%) $\uparrow$
& Precision (\%) $\uparrow$
& F1-Score(\%) $\uparrow$\\
\midrule

\multirow{4}{*}{R2R}

& DEX-AR~\cite{bousselham2026dex}
& Gradient 
&14.54 
&9.12 
&11.21\\

& GradCAM~\cite{selvaraju2017grad}
& Gradient 
&17.66 
&35.91 
&23.68\\

& TAM~\cite{li2025token}
& Logit 
&14.92
&7.13 
&9.65\\

& \textbf{Ours} 
& Gradient 
&\textbf{31.98} 
&\textbf{43.70} 
&\textbf{36.93}\\

\midrule

\multirow{4}{*}{RxR}

& DEX-AR~\cite{bousselham2026dex}
& Gradient 
&21.13 
&7.04
&10.56\\

& GradCAM~\cite{selvaraju2017grad}
& Gradient 
&27.83 
&28.62 
&28.22\\

& TAM~\cite{li2025token}
& Logit 
&23.32 
&6.24 
&9.85\\

& \textbf{Ours} 
& Gradient 
&\textbf{28.46} 
&\textbf{42.41} 
&\textbf{34.06}\\

\bottomrule
\end{tabular}
}
\end{table}

%% file: Items/tab_generalization.tex











\begin{table}[t]
\centering
\caption{
\textbf{Performance gains over DEX-AR across different VLN policies.}
All values denote improvements.
}
\label{tab:agent_results}

\small
\setlength{\tabcolsep}{3.2pt}
\renewcommand{\arraystretch}{1.05}

\begin{tabular}{@{}c|c|ccc@{}}
\toprule
\multirow{2}{*}{Dataset}
& \multirow{2}{*}{VLN Policy}
& \multicolumn{3}{c}{\textbf{Ours over DEX-AR~\cite{bousselham2026dex} }} \\
\cmidrule(l){3-5}
&
& $\Delta$NLL  (\%)  $\uparrow$
& Prec.  (\%)  $\uparrow$
& F1  (\%)  $\uparrow$ \\
\midrule

\multirow{3}{*}{{R2R}}
& NaVid~\cite{zhang2024navid}
& +14.23 & +18.23 & +16.50 \\

& NaVILA~\cite{cheng2024navila}
& \textbf{+17.44} & \textbf{+34.58} & \textbf{+25.72} \\

& StereoNav~\cite{wang2026limits}
& +5.16 & +34.31 & +9.17 \\

\midrule

\multirow{3}{*}{{RxR}}
& NaVid~\cite{zhang2024navid}
& +11.86 & +37.50 & +22.58 \\

& NaVILA~\cite{cheng2024navila}
& +7.33 & +35.37 & +23.50 \\

& StereoNav~\cite{wang2026limits}
& \textbf{+18.70} & \textbf{+42.03} & \textbf{+25.90} \\

\bottomrule
\end{tabular}
\end{table}

%% file: Items/fig_ablation_TR.tex
\begin{figure}[t]
    \centering
    \includegraphics[width=\columnwidth]{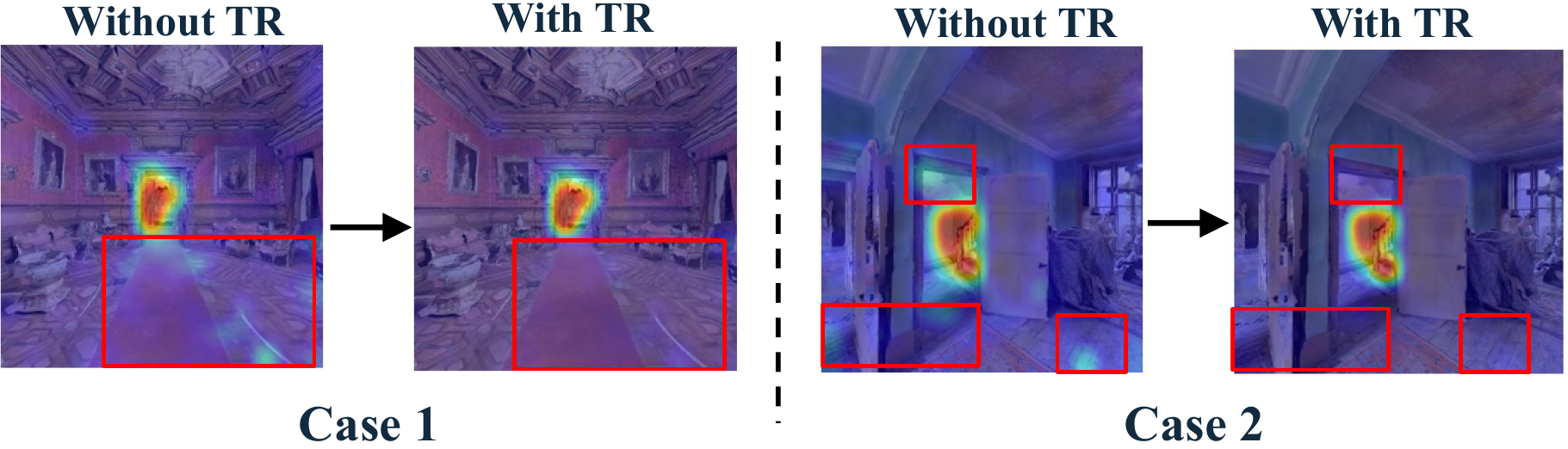}
    \vspace{-6mm}
    \caption{
    \textbf{Effect of TR.}
Comparisons before and after TR show that stage-guided historical support reduces cross-stage interference and yields more focused attribution on evidence relevant to the current navigation decision. Red boxes highlight the irrelevant regions.
    }
    \vspace{-2mm}
    \label{fig:temporal_refinement_ablation}
\end{figure}

%% file: Items/fig_ablation_SF.tex
\begin{figure}[t]
    \centering
    \includegraphics[width=\columnwidth]{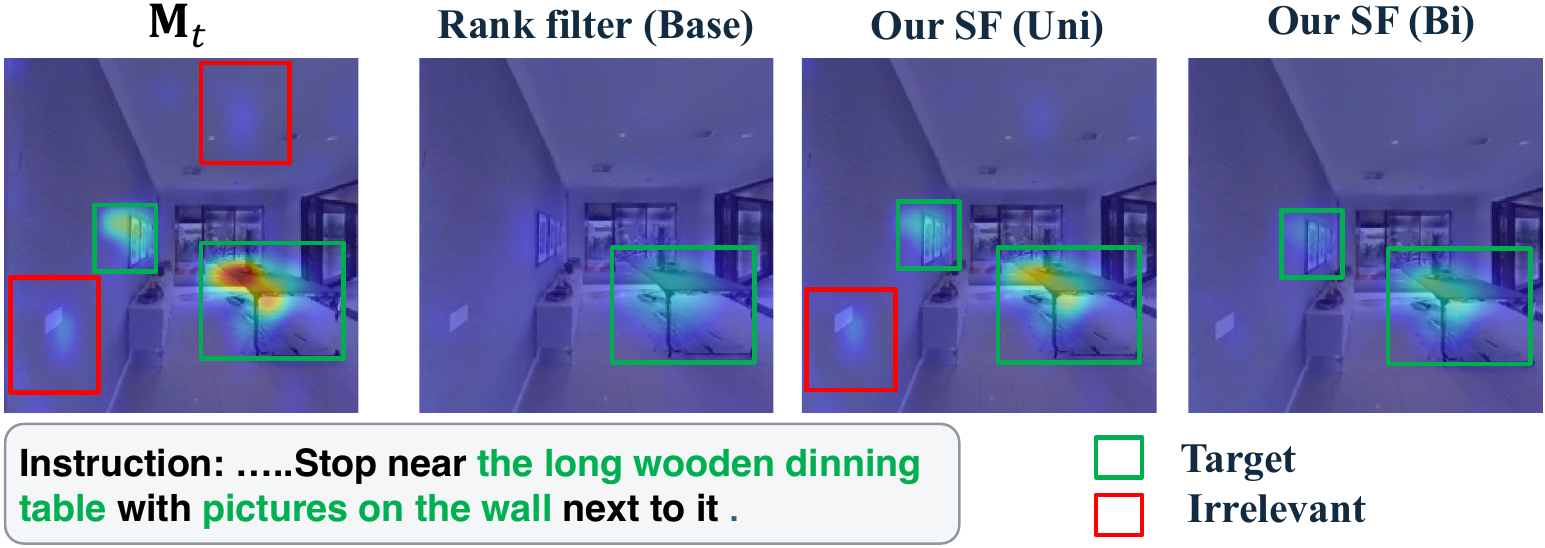}
    \vspace{-4mm}
    \caption{
    \textbf{Effect of SF.}
Visualization of the base rank filter~\cite{li2025token} and our uni-/bidirectional SF variants (Section~\ref{sec:spatial_filter}). Green and red boxes denote instruction-relevant and irrelevant regions, respectively.
    }
    \vspace{-4mm}
    \label{fig:semantic_filter_ablation}
\end{figure}

%% file: Items/tab_ablation.tex
\begin{table}[t]
\centering
\caption{\textbf{Ablation of temporal refinement (TR, Sec.~\ref{sec:temporal_refinement})
and spatial filtering (SF, Sec.~\ref{sec:spatial_filter}) on R2R.}}
\label{tab:ablation}
\setlength{\tabcolsep}{8pt}
\renewcommand{\arraystretch}{1.12}
\begin{tabular}{cc|ccc}
\toprule
\textbf{TR} & \textbf{SF}
& $\Delta\mathrm{NLL}(\%)\uparrow$
& \textbf{Precision}(\%)$\uparrow$
& \textbf{F1-Score}(\%)$\uparrow$ \\
\midrule
$\times$     & $\times$     
& 26.61 
& 41.80 
& 32.52 \\

$\checkmark$ & $\times$     
& 27.81 
& 43.28 
& 33.86 \\

$\times$     & $\checkmark$ 
& 30.82 
& 41.93 
& 35.53 \\

$\checkmark$ & $\checkmark$ 
& \textbf{31.98}
& \textbf{43.70} 
& \textbf{36.93} \\

\bottomrule
\end{tabular}
\end{table}

%% file: Items/fig_statics.tex
\begin{figure}[t]
    \centering
    \includegraphics[width=1\columnwidth]{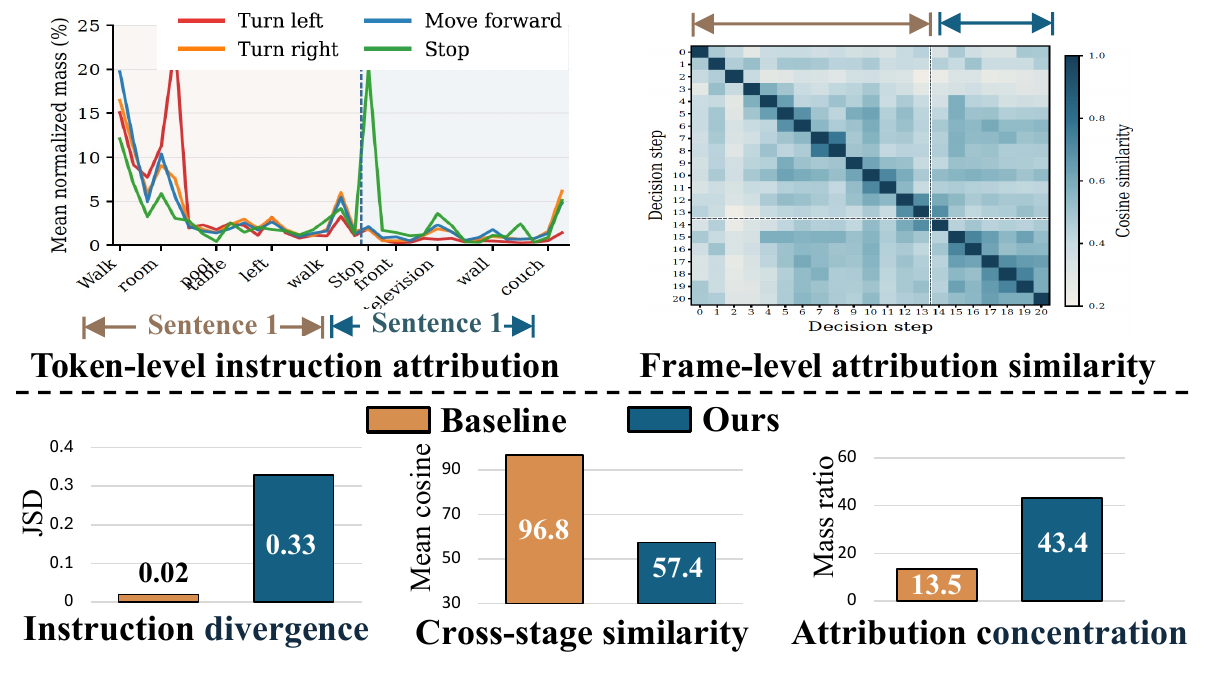}
    \vspace{-6mm}
   \caption{
   \textbf{Quantitative analysis of instruction-stage insensitivity.}
\textbf{Top:} token-level instruction attribution and frame-level similarity.
\textbf{Bottom:} dataset-level comparison with the baseline,~\ie, TAM~\cite{li2025token}.
Our method 
\textbf{\emph{shows stronger sensitivity
to instruction-stage transitions and more concentrated visual explanations.}}
}
\label{fig:statics}
\vspace{-3mm}
\end{figure}

%% file: Items/fig_real_world.tex
\begin{figure*}[t]
    \centering
    \includegraphics[width=2\columnwidth]{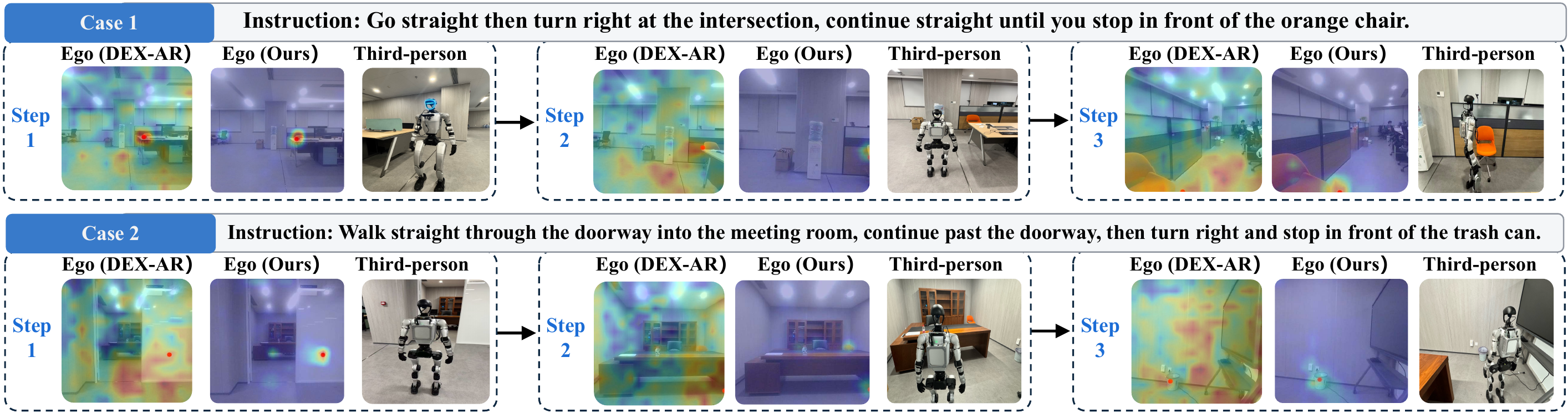}
    \caption{
   \textbf{Real-world qualitative comparison on a Unitree G1 humanoid robot.}
Two representative navigation cases are shown across multiple decision steps.
For each step, we compare egocentric attribution maps from DEX-AR~\cite{bousselham2026dex} and our method,
together with the corresponding third-person execution view.
Our method better follows the evolving instruction stage and highlights
decision-relevant visual evidence throughout real-world navigation
}
    \label{fig:unitree}
    \vspace{-3mm}
\end{figure*}

%% file: Items/fig_failure_analyze.tex
\begin{figure}[!t]
    \centering
    \includegraphics[width=\columnwidth]{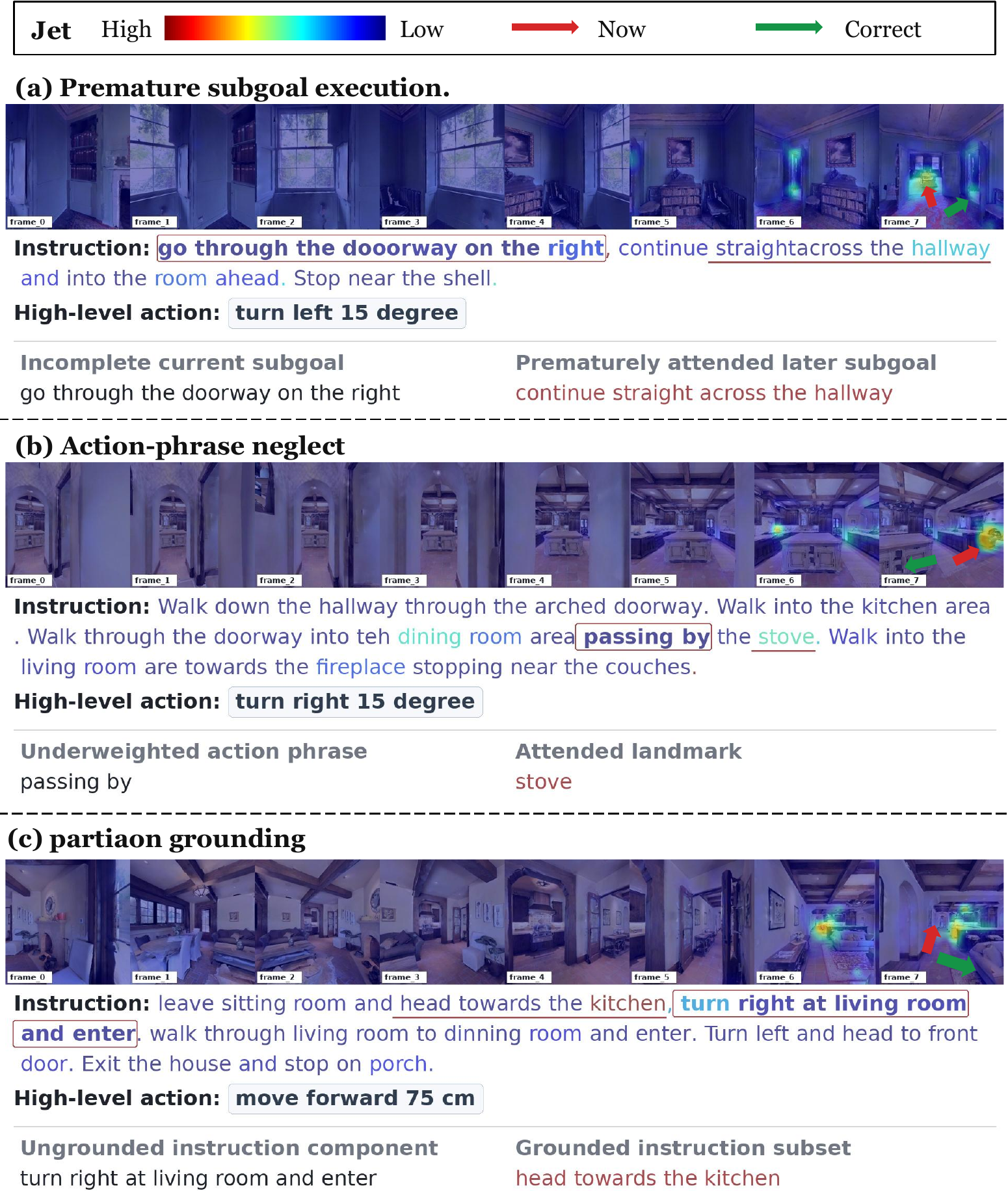}
    \caption{
  \textbf{Our method as a diagnostic tool for VLN policies.}
Our attributions reveal stage- and token-level causes of VLN failures.
\textbf{(a) Premature subgoal execution}: attention shifts to a later subgoal too early.
\textbf{(b) Action-phrase neglect}: the landmark is emphasized while its action phrase is underweighted.
\textbf{(c) Partial instruction grounding}: only part of the relevant instruction is grounded.
    }
    \vspace{-4mm}
    \label{fig:failure}
\end{figure}

%% file: Secs/5_conclusion.tex
\section{Conclusion}

We presented a stage-aware attribution framework for VLN that captures
instruction--visual evidence across navigation history and adapts explanations
to changing instruction stages. Experiments on R2R and RxR show improved
semantic alignment while maintaining strong decision faithfulness, with further
applications to failure diagnosis and real-world navigation on a Unitree G1
robot. These results highlight the value of stage-aware explanation for
long-horizon VLN. Future work will extend the framework to longer histories
and continuous low-level control.